\pdfoutput=1
\documentclass[11pt,a4paper]{article}
\usepackage[hyperref]{acl2021}
\usepackage{times}
\usepackage{latexsym}
\usepackage{xcolor}
\usepackage[normalem]{ulem}
\usepackage{graphicx}
\usepackage{footnote}
\makesavenoteenv{tabular}
\makesavenoteenv{table}

\usepackage{soul}

\usepackage{microtype}

\aclfinalcopy % Uncomment this line for the final submission
\def\aclpaperid{2891} %  Enter the acl Paper ID here

\newcommand{\bs}[1]{{\textcolor{black}{#1}}{\bf}}
 \newcommand{\lb}[1]{{\textcolor{black}{#1}}}

\newcommand{\mg}[1]{{\textcolor{black}{#1}}}

\title{\textit{How to} Split: the Effect of Word Segmentation on Gender Bias\\ in Speech Translation}

\author{Marco Gaido\textsuperscript{1,2 $\dagger$}, Beatrice Savoldi\textsuperscript{1,2 $\dagger$}, Luisa Bentivogli\textsuperscript{1}, Matteo Negri\textsuperscript{1}, Marco Turchi\textsuperscript{1} \\
  \textsuperscript{1}Fondazione Bruno Kessler, Trento, Italy \\
  \textsuperscript{2}University of Trento, Italy \\
  \texttt{\{mgaido,bsavoldi,bentivo,negri,turchi\}@fbk.eu} 
  }

\date{}

\begin{document}
\maketitle
\begin{abstract}
\newcommand\blfootnote[1]{%
  \begingroup
  \renewcommand\thefootnote{}\footnote{#1}%
  \addtocounter{footnote}{-1}%
  \endgroup
}
\blfootnote{\textsuperscript{$\dagger$}The authors contributed equally.}
Having recognized gender bias as a major issue affecting current translation technologies, researchers have primarily attempted to mitigate it by working on the data front. However, whether algorithmic aspects concur to exacerbate unwanted outputs remains so far under-investigated. In this work, we bring the analysis on gender bias in automatic translation onto a seemingly neutral yet critical component: word segmentation. Can segmenting methods influence the ability to translate gender? Do certain segmentation approaches penalize the representation of feminine linguistic markings? We address these questions by comparing 5 existing segmentation strategies on the target side of speech translation systems. Our results on two language pairs (English-Italian/French) show that state-of-the-art subword splitting (BPE) comes at the cost of higher gender bias. In light of this finding, we propose a combined approach that preserves BPE overall translation quality, while leveraging the higher ability of character-based segmentation to properly translate gender.

\end{abstract}

\paragraph{Bias Statement.\footnote{As suggested by \cite{blodgett-etal-2020-language} and required 
for other venues \bs{\cite{hardmeier2021write},}
%(\url{https://genderbiasnlp.talp.cat/}), 
we formulate our bias statement.
}}
We study the effect of segmentation methods on the ability of speech translation (ST) systems to translate masculine and feminine forms referring to  
human entities. 
In this area, structural linguistic properties interact with the perception and representation of 
individuals 
%men and women
\cite{Gygaxindex4, corbett2013expression, stahlberg2007representation}. 
Thus, we believe 
they are relevant gender expressions, used to communicate about the self and others, and by which  the sociocultural and political reality of gender is negotiated \cite{bookgender}. 

Accordingly, we consider a model that systematically and disproportionately favours masculine over feminine forms to be biased, as it fails to properly recognize women. 
From a technical perspective, such behaviour deteriorates 
models' performance. Most importantly, however, from a human-centered view, 
%there are 
real-world 
harms are at stake \cite{crawford2017trouble}, as
translation technologies
are unequally beneficial across gender groups
and reduce feminine visibility, thus contributing
to misrepresent 
an already 
socially 
disadvantaged group. 

This work is motivated by the intent to  shed light on whether 
issues in the generation of feminine forms are
also a by-product of current algorithms and techniques.
In our view, architectural improvements of 
ST systems should 
also 
account for the trade-offs between  overall translation quality and
gender representation:
our proposal of a model that combines two segmentation techniques
is a step towards this goal.

Note that technical 
mitigation approaches 
should be integrated with the
long-term multidisciplinary commitment \cite{criado2019invisible,Ruha,d2020data} necessary to radically address bias in our community. 
Also, we recognize the limits of working on 
binary gender, as we further discuss in the ethic 
section
({$\S$}\ref{sect:ethic}).

\section{Introduction}

The widespread use of language technologies has motivated growing interest on their social impact \cite{hovy-spruit-2016-social,blodgett-etal-2020-language}, 
with  gender bias  
%and other demographic 
%\cite{kiritchenko-mohammad-2018-examining,sap-etal-2019-risk} 
representing a major cause of concern \cite{costa-nature, sun-etal-2019-mitigating}.
As regards translation tools, focused evaluations have exposed that
speech translation (ST) -- and machine translation (MT) -- models
do in fact 
overproduce masculine references in their outputs \cite{cho-etal-2019-measuring,bentivogli-2020-gender}, except for
feminine associations perpetuating traditional gender roles and 
stereotypes
\cite{Prates2018AssessingGB, stanovsky-etal-2019-evaluating}.
In this context, most works 
identified \textit{data} as the primary source 
of
gender asymmetries. 
Accordingly,
many pointed out the 
 misrepresentation  of gender groups in datasets  \cite{broadcast, vanmassenhove-etal-2018-getting}, focusing on the development of data-centred mitigating techniques
 \cite{zmigrod-etal-2019-counterfactual,saunders-byrne-2020-addressing}.
 
Although data 
are 
not the only
factor contributing to generate bias \bs{\cite{shah-etal-2020-predictive,savoldi2021gender}}, 
only few inquiries devoted attention to other \bs{technical}
components that
exacerbate the problem \cite{vanmassenhove-etal-2019-lost} 
or to architectural changes that can
contribute to its mitigation \cite{costajussa2020gender}.
From an algorithmic perspective, \citet{roberts2020decoding} additionally expose how ``taken-for-granted''  approaches
may come with high overall translation quality in terms of BLEU scores, but are actually detrimental when it comes to gender bias.

Along  this line,  we focus on ST systems and inspect a core  aspect of neural %MT
models: word segmentation. Byte-Pair Encoding (BPE) \cite{sennrich-etal-2016-neural} represents the
\textit{de-facto} standard and has recently shown to yield
better results compared to character-based segmentation
in ST \cite{di-gangi-2020-target}. But does this hold true for gender translation as well? If not, why?

%Languages like
%%Italian and French
%\mg{French and Italian} often exhibit comparatively complex feminine forms, derived from the masculine ones by means of an additional suffix (e.g. en: \textit{professor}, fr: \textit{professeur} M vs. \textit{professeur\underline{e}} F). In light of the above, purely statistical segmentation methods could be unfavourable for gender translation, as they can break the morphological structure of words and thus lose relevant linguistic information \cite{ataman-2017-lmvr}. Additionally, women and their referential linguistic expressions of gender are typically under-represented in existing corpora \cite{hovy-etal-2020-sound}. Consequently, as BPE merges the character sequences that co-occur more frequently, rarer (and more complex) feminine-marked words may result in less compact sequences of tokens (e.g. en: \textit{described}, it:  \textit{des@@critt\underline{o}} M vs. \textit{des@@crit@@t\underline{a}} F). In light of such typological and distributive conditions, may certain splitting methods render feminine gender less probable and hinder its prediction?
Languages like French and Italian often exhibit comparatively complex feminine forms, derived from the masculine ones by means of an additional suffix (e.g. en: \textit{professor}, fr: \textit{professeur} M vs. \textit{professeur\underline{e}} F). Additionally, women and their referential linguistic expressions of gender are typically under-represented in existing corpora \cite{hovy-etal-2020-sound}. In light of the above, purely statistical segmentation methods could be unfavourable for gender translation, as they can break the morphological structure of words and thus lose relevant linguistic information \cite{ataman-2017-lmvr}. Indeed, as BPE merges the character sequences that co-occur more frequently, rarer or more complex feminine-marked words may result in less compact sequences of tokens (e.g. en: \textit{described}, it:  \textit{des@@critt\underline{o}} M vs. \textit{des@@crit@@t\underline{a}} F). Due to  such typological and distributive conditions, may certain splitting methods render feminine gender less probable and hinder its prediction?

We address such questions by implementing different families of segmentation approaches employed on the decoder side of ST 
models built on the same training data. By comparing the resulting models both in terms of overall translation quality and gender accuracy, we explore whether a so far considered irrelevant  aspect like  word segmentation 
%\mg{(cfr. \citealt{bentivogli-2020-gender})} 
can  actually
affect gender translation.
As such,
  \textbf{(1)}
  we perform the first comprehensive analysis of the results obtained by 5 popular segmentation techniques for two language directions
(en-fr and en-it) in ST.
\textbf{(2)} We find that 
the target segmentation method is indeed an important factor for models' gender bias.  Our experiments  consistently show that
BPE leads to the highest BLEU scores, while character-based models are the best at translating gender.
%Further analysis
\bs{Preliminary analyses}
suggests that the
isolation of the morphemes encoding gender 
%is
\bs{can be}
a key factor for gender translation. 
\textbf{(3)} 
Finally, we propose a
multi-decoder architecture able to  combine BPE overall translation quality and the higher  ability  to  translate  gender  of character-based segmentation. 

\section{Background}

\paragraph{Gender bias.}
Recent years have seen a surge of studies dedicated to gender bias in 
MT
\cite{gonen-webster-2020-automatically, rescigno-etal-2020-case}
and ST
%\cite{bentivogli-2020-gender}
\cite{costa2020evaluating}. 
The primary source of such gender imbalance and adverse outputs
has been identified in the
training data, which reflect the under-participation of women  
-- e.g. in the media
%\cite{garnerin-etal-2020-gender, 
\cite{pmlr-v81-madaan18a}, sexist language and gender categories overgeneralization \cite{devinney-etal-2020-semi}.
Hence, preventive initiatives concerning data documentation have emerged \cite{bender-friedman-2018-data}, and several mitigating strategies have been proposed by
training models on \textit{ad-hoc}
gender-balanced datasets \cite{saunders-byrne-2020-addressing, costa-jussa-de-jorge-2020-fine}, 
or by
enriching
data with additional gender information 
\cite{moryossef-etal-2019-filling,vanmassenhove-etal-2018-getting, elaraby-zahran-2019-character, saunders-etal-2020-neural, stafanovics-etal-2020-mitigating}.
%, gaido-etal-2020-breeding}.

Comparatively, very little work has tried to identify concurring factors to gender bias going beyond data. Among those,
\citet{vanmassenhove-etal-2019-lost}
ascribes to an algorithmic bias the loss of less frequent feminine forms in both phrase-based and neural MT.
 Closer to our intent,
 two recent works pinpoint the impact of models' components and inner 
 mechanisms. 
 \citet{costajussa2020gender} investigate the role of different architectural designs in
 multilingual MT, showing that
 language-specific encoder-decoders \cite{escolano-etal-2019-bilingual}
 better translate gender than shared models
 \cite{johnson-etal-2017-googles}, as the former retain more gender information in the source embeddings and keep more diversion in the attention.
 \citet{roberts2020decoding}, on the other hand, 
 prove that
 the adoption of beam search instead of sampling -- although beneficial in terms of BLEU scores -- has an impact on gender bias. Indeed, it leads models to an extreme operating point that exhibits zero variability and in which they tend to generate the more frequent (masculine) pronouns.
Such
studies therefore expose largely unconsidered aspects
as factors contributing to gender bias in automatic translation,
identifying future research directions for the needed countermeasures. 

To the best of our knowledge, no prior work has
%so far
taken into account if it may be the case for segmentation methods as well. Rather, prior work in ST \cite{bentivogli-2020-gender}
%has 
compared gender translation performance of cascade 
%vs.
and
%end-to-end
direct systems
%trained with the use of different compression
using different segmentation algorithms, 
%assuming it would not affect their results. 
disregarding their possible 
%influence
impact on final results.
% Closest to our intent are the works by (Costa Jussà et al, 2020) - which investigates the role different architectural designs play in gender bias in Multilingual MT - and (Roberts et al, 2020) proving that beam search is comparatively with sampling more biased toward the generation of more frequent (masculine) pronouns. In both works, it has been shown while “taken-for-granted” approaches, like -respectively - the use of Transformer architecture or beam search, improve overall translation quality in terms of BLEU scores, they are disadvantaged when it comes to gender bias. 
% Similarly, the BPE compression method is the MT state-of-the-art approach, and has been recently proved to be the highest-quality segmentation technique in ST as well (Di Gangi et al, 2020). But does it hold true for gender translation as well?

\paragraph{Segmentation.}
%Due to the vocabulary coverage problem, the segmentation of words into smaller atomic elements is a critical component of 
%NMT
%neural models.
%\mg{The need to deal with unseen words at inference time has been addressed by open-vocabulary methods that segment words into smaller atomic elements.}
Although early attempts in neural MT employed word-level sequences \citep{sutskever-2014-nmt,bahdanau-2015-nmt},
the need for open-vocabulary systems able to translate rare/unseen words led to the definition of
several 
%subword
word segmentation techniques.
Currently, the statistically motivated approach based on 
%the 
byte-pair encoding (BPE) %model proposed 
by  \citet{sennrich-etal-2016-neural} represents the \textit{de facto} standard in MT. Recently, its superiority to character-level \citep{costa-jussa-2016-character, chung-etal-2016-character} has been also proved in the context of ST 
%for eight language pairs
\citep{di-gangi-2020-target}. However, 
%prior work \cite{} has shown that 
depending on the languages involved in the translation task, the data conditions, and the linguistic properties taken into account, BPE greedy 
%procedure
procedures
%exhibit
can be suboptimal. By breaking the surface of words into plausible semantic units, linguistically motivated segmentations \citep{smit-etal-2014-morfessor,ataman-2017-lmvr} 
%have been 
were
proven more effective for low-resource and morphologically-rich languages (\bs{e.g. agglutinative languages like Turkish}), which often have a high level of sparsity in the lexical distribution due to their numerous derivational and inflectional variants. Moreover, fine-grained analyses comparing the grammaticality of character, morpheme and BPE-based models %have 
exhibited different capabilities.
%several compromises.
%trade-offs.
% Accordingly,
\citet{sennrich-2017-how-grammatical} and \citet{ataman-etal-2019-importance} show the syntactic advantage of BPE 
%over character-sequences 
in managing several agreement phenomena in German, a language that requires resolving long range dependencies. 
%On the other side,
In contrast, \citet{Belinkov2019OnTL}
%probe
demonstrate that while
subword units better capture
%\mg{BPE better captures} 
semantic information, character-level representations perform best at generalizing morphology, thus being more robust 
%towards 
%\bs{for}
in handling unknown and low-frequency words. Indeed, using different atomic units does affect models' ability
%the representation of different atomic units do affect translation models and their ability 
to handle specific linguistic phenomena. However, whether low gender translation accuracy can be to a certain extent considered a by-product of certain compression algorithms is 
%yet 
still
unknown.

\section{Language Data}
\label{sec:length}
\bs{As just discussed, the effect of segmentation strategies can vary depending on language typology \cite{ponti2019modeling} and data conditions.}
To inspect the interaction between word segmentation and gender expressions, we \bs{thus} first clarify the properties of grammatical gender in the two languages of our interest: French and Italian. Then, we verify their representation in 
%\bs{the corpus used for our experiments in ST (see $\S$\ref{sec:experiments}).} 
the %evaluation set used 
datasets used for our experiments.
%(see $\S$\ref{sec:evaluation}).
%\bs{in ($\S$\ref{sec:experiments}).}
%\ref{sec:evaluation}).
%the MuST-SHE corpus \cite{bentivogli-2020-gender}, which we use in our ST evaluations  (see $\S$\ref{sec:evaluation}).

% \subsection{Gender and ?}
\subsection{Languages and Gender}
\label{sec:language}
% \bs{what abut this section? Fix, expand, (re)move?}
The extent to which
information about
the gender of referents is grammatically encoded varies across languages \cite{bookgender,Gygaxindex4}. Unlike English -- whose gender distinction is chiefly displayed via pronouns 
%in pronominal forms 
(e.g. \textit{he/she}) -- fully grammatical gendered languages like French and  Italian systematically articulate such semantic distinction 
%(gender assignment) 
on several parts of speech (gender agreement) \cite{Hockett:58, Corbett:91}.  
%exhibit extensive structural system of semantic (gender assignment) and morpho-syntactic (gender agreement) features on numerous parts of speech \cite{Hockett:58,Corbett:91}. 
Accordingly, 
%this results in 
many lexical items 
%(nouns, verbs, adjectives and determiners%, too) that exists 
exist
in both feminine and masculine variants\mg{,}
%\footnote{Note that, in some cases, determiners alone can alone be responsible for overt markings e.g. en: \textit{the musician}, it: \underlile{il} vs. \underlile{la} \textit{musicista.}} 
overtly marked through morphology (e.g. en: \textit{the tired kid sat down}; it: \textit {\underline{il} bimb\underline{o} stanc\underline{o} si \`e sedut\underline{o}} M vs. \textit{\underline{la} bimb\underline{a} stanc\underline{a} si \`e  sedut\underline{a}} F).
%
%it: \textit{\underline{la} bimb\underline{a} stanc\underline{a} si \`e  sedut\underline{a}} vs. \textit{\underline{il} bimb\underline{o} stanc\underline{o} si \`e sedut\underline{o}}). }
%
%\mt{Secondo me qui non ci vuole il new line, perche' commenti l'esempio sopra}
As the example shows, the word forms are distinguished by two morphemes ( \textit{--o}, \textit{--a}), which respectively represent the most common
%\footnote{There are due exceptions, e.g. the word \textit{musicist\underline{a} (\textit{musician}}) is invariable for both masculine/feminine references.} 
%%%SPAZIO Also, words with ''\textit{-e}'' ending are often epicene (e.g. en: ``\textit{the happy kid}'', it: \textit{\underlile{il}/\underline{la} bimb\underline{o/a}} felice).}} 
%
%With due exceptions,e.g. contrastive nominal forms of \textit{musician} and \textit{singer} do not exist (\textit{il/la cantant\underline{e}}; \textit{il/la musicist\underline{a}}.} 
inflections for Italian masculine and feminine markings.\footnote{In a fusional language like Italian, one single morpheme can denote several properties as, in this case, gender and singular 
% number. The plural forms would be \textit{bimb}\underline{e} vs. \textit{bimb}\underline{i}.} 
number (the plural forms would be \textit{bimb}\underline{i} vs. \textit{bimb}\underline{e}).}
%
%in Italian two different morphemes ( \textit{--a}, \textit{--o}) carry feminine and masculine markings, respectively. 
%
%
In French, the morphological mechanism is slightly different~\citep{Schaforth}, as it %commonly 
relies on an additive suffixation on top of masculine words
%for marking
to express feminine gender  (e.g. en: \textit{an expert is gone}, fr: \textit{un expert est allé} M vs. \textit{un\underline{e} expert\underline{e} est allé\underline{e}} F).
%\textit{un\underline{e} expert\underline{e} est allé\underline{e}} vs. \textit{un expert est allé}).}
%with plural often being further inserted with an  \textit{s} (e.g. expert\underline{es} vs expert\underline{s}. 
Hence, feminine French forms require an additional morpheme.
%while the masculine ones also carry ttense and person
Similarly, another productive %feminization 
strategy --
%\bs{cit. Gyarg}, 
typical for a %sub
set of personal nouns -- is the derivation of feminine words 
%with use of
via
specific affixes for both French (\textit{e.g. --eure, --euse})\footnote{French also requires additional modification on feminine forms due to phonological rules (e.g. en: \textit{chef/spy}, fr: \textit{chef\underline{fe}}/\textit{espion\underline{ne}} vs. \textit{chef}/\textit{espion}).} and Italian (\textit{--essa, --ina, --trice}) \cite{Schaforth, chini1995genere}, 
%a strategy 
whose residual evidence is still 
found
%present 
in some English forms (e.g. \textit{\bs{heroine}, actress}) \cite{umera2012linguistic}. 
%\bs{this are trends, but there is no assurance that --e in French or --a in Italian necessarily refere to semantics of feminine (e.g. poeta = masculine), plus there are many syncretism and plus the same form are also used on formally feminine inanimate nouns.}

In light of the above, translating gender from English into French and  Italian 
%phenomena 
poses several challenges to automatic
%ST 
models. 
%For ST, such phenomenon poses several challenges. 
First, gender translation %from English into Italian and French 
does not allow for
%words
one-to-one mapping between source and target words. Second,
%due to the richer morphology of the target languages,
%%there are more issues of data sparsity as the model should learn more word variants.
%the greater number of variants increases data sparsity.
the richer morphology of the target languages increases the number of variants and thus data sparsity.
Hereby, the question is whether --  and to what extent -- statistical word segmentation  differently treats the less 
%\bs{distributed} 
{frequent}
%ones
variants. Also, 
%due to higher level of 
considering the morphological complexity of some feminine forms, we speculate whether linguistically
%inconsiderate
unaware splitting may
%lose relevant information and 
disadvantage their translation.
%\bs{generation} 
%representation. %of feminine forms. 
To test these hypotheses, below we %first 
%verify that 
explore if such conditions are represented in
% %our
% the used  ST 
% %evaluation 
% datasets.
the ST datasets used in our study.

%\bs{Fusional languages have the tendency to tendency to use a single inflectional morpheme to denote multiple grammatical, syntactic, orsemantic features. On the other hand, in agglutinative languages, each morpheme in a word remains in every aspect unchanged after their composition, allowing a direct identification of the morpheme boundaries. In fusional and agglutinating typologies, morphemes are generally composed continuously to construct new word forms. On the other hand, it is also possible to observe templatic typologies, for instance in Arabic, where morphemes are inserted in certain templates in a discontinuous fashion to achieve certain derivative or inflective transformations. }

%\subsubsection{\bs{Used Corpora and Gender \lb{Representation}}}
\subsection{Gender 
%Representation
in Used Datasets}
%Corpora
% \bs{For our experiments, we rely on the... } 
%\subsection{MuST-SHE: Preliminary Analysis}
\label{sec:preliminary}
MuST-SHE \cite{bentivogli-2020-gender} is a gender-sensitive benchmark available for both en-fr and en-it
%\lb{\footnote{\url{ict.fbk.eu/must-she/}}}  
%(1,113 sentences for en-fr and 1,096 for en-it). 
(1,113 and 1,096 sentences, respectively). 
Built on naturally occurring instances of gender phenomena retrieved from the TED-based MuST-C corpus \cite{MuST-Cjournal},\footnote{Further details about these datasets are provided in {$\S$}\ref{sect:ethic}.} it allows to evaluate gender translation on qualitatively differentiated and %various realization of 
%equally 
balanced masculine/feminine forms. An important feature of MuST-SHE is that, for each reference translation, an almost identical ``wrong'' reference is created by swapping each annotated gender-marked word into its opposite gender. By means of such wrong reference, 
%per
for
each target language we 
can 
%thus 
identify 
%Such wrong reference thus permits to identify 
$\sim$2,000 pairs of gender forms  (e.g. en: \textit{tired}, fr: \textit{fatigué\underline{e}} vs. \textit{fatigué})
%,
%, on which we rely to 
that we compare in terms of \textit{i)} 
length, and \textit{ii)} frequency in the MuST-C training set.

%%%%% ORIG
% %the frequency of both word forms 
% %\bs{their frequency distribution}
% their frequency in the MuST-C
% %set used for training;
% training set, and
% %, the %TED-based  corpus used to train our models
% %;
% \textit{ii)} the length of their character sequences.   
%%%%%%%%
As regards frequency, 
we asses that, 
for both language pairs, \bs{the types of} feminine variants 
%results
are less frequent than their masculine counterpart in over 86\% of the cases. 
Among the exceptions, we find words that are almost gender-exclusive (e.g. \textit{pregnant}) and some problematic or socially connoted activities (e.g. \textit{raped, nurses}). 
% married}).
%
%Looking at sequence length, MuST-SHE reflects the typological features described in {$\S$}\ref{sec:language}. In fact, 15\% of Italian feminine forms result in longer sequences, whereas
%mg{in French this percentage amounts to almost 95\%.}
Looking at 
%sequence 
words' length, 15\% of Italian feminine forms result to be longer than masculine ones, 
%in longer  sequences, 
whereas in French this percentage amounts to almost 95\%. These scores  confirm that MuST-SHE reflects the typological features described in {$\S$}\ref{sec:language}.

%QUESTI RISULTATI CI PERMETTONO DI CONFERMARE CHE IL PROBLEMA DELLA TRADUZIONE DEL GENERE PUO’ ESSERE INQUADRATO COME UN PROBLEMA LEGATO ALLA LUNGHEZZA DELLE SEQUENZE DI PAROLE A FORMA FEMMINILE E DI PAROLE MENO FREQUENTI. 

\section{Experiments}
\label{sec:experiments}
%\bs{
All the direct ST systems used in our experiments are built in the same fashion within a controlled environment, so to keep the effect of different word segmentations as the only variable. Accordingly, we train them on the 
%fairly uniform 
MuST-C corpus, which contains 492 hours of speech for en-fr and 465 %hours 
for en-it.
%}
%
%As per
Concerning the architecture, our models are based on  Transformer \cite{transformer}. For the sake of reproducibility, we provide extensive details about the ST models and hyper-parameters' choices in the
\mg{Appendix $\S\ref{appsec:models}$}.\footnote{\mg{Source code available at \url{https://github.com/mgaido91/FBK-fairseq-ST/tree/acl_2021}.}}
%supplementary materials.

%
%Our direct ST models used in the experiments of this paper are based on the Transformer \cite{transformer} and are described extensively in the Appendix. They were trained on the MuST-C corpus \cite{MuST-Cjournal}, which contains 492 hours of speech for en-fr and 465 hours for en-it.
%
\subsection{Segmentation Techniques}
%
%Although early attempts in neural MT employed word-level sequences \cite{bahdanau-2015-nmt,sutskever-2014-nmt},
%the need for open-vocabulary systems able to translate rare/unseen words led to the definition of
%several subword segmentation techniques.
%In particular,
%\mg{In this work,} we identified three different 
%types 
%of segmentation techniques. %proposed over the years.
To allow for a comprehensive comparison of word segmentation's impact on gender bias in ST, we identified three substantially different categories 
%types 
of splitting techniques. 
For each of them, we hereby present the candidates selected for our experiments.  
%For each of them, we select one (or more) candidates that we evaluate in our experimentsto compare their effect on gender bias.

\paragraph{Character Segmentation.}
Dissecting words at their 
%minimal 
maximal level of granularity, 
%As statistical methods pose a bias towards more frequent subwords,
\textbf{character-based} solutions have been first proposed by
\citet{ling2015characterbased} and \citet{costa-jussa-2016-character}.
%Character-level segmentation is
%With words treated as sequences of characters, this
This technique proves
simple and particularly effective at generalizing over unseen words. 
% simple and particularly effective at learning morphology and rendering rare words compared to BPE \cite{Belinkov2019OnTL},
%but struggles in the presence of long-range dependencies \cite{sennrich-2017-how-grammatical}.
%.
On the other hand, 
%However, 
the 
%increased 
length of
%such
the resulting
sequences
%requires character-based models to have more memory, thus slowing them down both at training and inference time.
increases the memory footprint, 
%slowing both 
and slows both the
training and inference phases.  
%Besides, character-based models are slower both at training and inference time and require more memory, as the resulting sequences are significantly longer.
We perform our segmentation 
%\mg{Our segmentation is performed
by appending ``@@ '' to all characters but the last of each word.
%\mn{IN PARALLELO CON LA FINE DEL PARAGRAFO PRECEDENTE (We trained the DPE segmentation...), AGGIUNGERE 2 PAROLE SU COME LA FACCIAMO.}

\paragraph{Statistical Segmentation.}
This family comprises data-driven algorithms 
% that segment words into sub-units based on their occurrences in training corpora.
that generate statistically significant subwords units. The most popular  
%instance of this type
one
is \textbf{BPE}
%)
\cite{sennrich-etal-2016-neural},\footnote{We use
%the
SentencePiece
%implementation
\cite{kudo-richardson-2018-sentencepiece}: \url{https://github.com/google/sentencepiece}.} which proceeds by merging the most frequently co-occurring characters or character sequences.
Recently, \citet{he2020-dynamic} introduced the Dynamic Programming Encoding \textbf{(DPE)} algorithm, which 
%performed
performs
competitively and was claimed to accidentally produce more linguistically-plausible subwords with respect to BPE.
DPE is obtained by training a mixed character-subword model.
As such, the computational cost of a DPE-based ST model is
%nearly
around twice that of a BPE-based one.
We trained the DPE segmentation
on the transcripts
and the target translations
of the MuST-C training set, using
the same 
settings of the original paper.\footnote{See \url{https://github.com/xlhex/dpe}.}

\paragraph{Morphological Segmentation.}
%Whilst remaining at subword level, a
%Linguistically-guided tokenization  following morpheme boundaries is a third possibility. 
A third possibility is linguistically-guided tokenization that follows morpheme boundaries. 
%The last family contains methods that segment words according to morphological properties.
%The morphological 
%
%%%TOLTA PER SPAZIO
%\mn{Morphological} segmentation was already proposed for \mn{phrase-based} MT \cite{virpioja-2007-morphology-aware} and has been investigated in the context of neural MT as well \cite{sanchez-cartagena-2016-abu}.
%
% One of the most widespread tools is \textbf{Morfessor} \cite{creutz2005unsupervised}. \cite{ataman-2017-lmvr} extended it to control the size of the output vocabulary and improve translation quality, giving birth to the \textbf{LMVR} segmentation method.
Among the unsupervised approaches, 
one of the most widespread tools is \textbf{Morfessor} \cite{creutz2005unsupervised}, which was extended by \citet{ataman-2017-lmvr} to control the size of the output vocabulary, giving birth to the \textbf{LMVR} segmentation method.
These 
%linguistically motivated segmentation 
techniques have 
%been shown to 
outperformed other approaches 
%in agglutinative and 
when dealing with low-resource and/or morphologically-rich languages \cite{ataman-2018-evaluation}. In other languages, they are not as effective, so 
they are not widely adopted.
Both Morfessor and LMVR have been trained on the MuST-C training set.\footnote{We used the parameters and commands suggested in \url{https://github.com/d-ataman/lmvr/blob/master/examples/example-train-segment.sh}}
%\mn{IN PARALLELO CON LA FINE DEL PARAGRAFO PRECEDENTE (We trained the DPE segmentation...), AGGIUNGERE 2 PAROLE SU COME LA FACCIAMO.}

% \begin{table}[h]
% %\setlength{\tabcolsep}{4pt}
% \centering
% \small
% \begin{tabular}{l|cc}
% & \textbf{en-fr} & \textbf{en-it} \\
% \hline
% BPE         & 8,048  & 8,064 \\
% Char        & 304   & 256 \\
% DPE         & 7,864  & 8,008 \\
% Morfessor   & 26,728 & 24,048 \\
% LMVR        & 21,632 & 19,264 \\
% \end{tabular}
% \caption{Resulting dictionary sizes.}
%   \label{tab:dictsize}
% \end{table}

%To have a 

\begin{table}[h]
\centering
\small
\begin{tabular}{l|cc}
& \textbf{en-fr} & \textbf{en-it} \\
\hline
\# tokens\footnote{
%Note that in the table 
Here ``tokens'' refers to the 
%total 
number of words in the corpus, and not to the unit resulting from subword tokenization.}  & 5.4M & 4.6M \\
\# types  & 96K & 118K \\
\hline
BPE         & 8,048  & 8,064 \\
Char        & 304   & 256 \\
DPE         & 7,864  & 8,008 \\
Morfessor   & 26,728 & 24,048 \\
LMVR        & 21,632 & 19,264 \\
\end{tabular}
\caption{Resulting dictionary sizes.}
%, as used in the paper

  \label{tab:dictsize}
\end{table}

For fair comparison, we chose the 
optimal
%best 
vocabulary size for each method (when applicable).
%As per 
Following \cite{di-gangi-2020-target}, we employed 8k merge rules for BPE and DPE, since the latter requires an initial BPE segmentation.
%For BPE, we used 8k merge rules, as the best size according to \cite{di-gangi-2020-target}, so we did for DPE that requires an initial BPE segmentation.
In LMVR, instead, the desired target dimension 
%size
is actually only an upper bound for the vocabulary size.
%that is not guaranteed to be respected.
We tested 32k and 16k, but we only report the results with
%as target vocabulary sizes. However,
 %we report only the results 
 %of
 32k
 %%%SPAZIO%%%%%
 %\footnote{\bs{As per preliminary analysis carried out with the Spacy  analyzer \cite{spacy}, the 32k bound was  ideal with respect to the distribution of morphological tags in MuST-C.}} 
 as it proved to be the best configuration both in terms of translation quality and gender accuracy.
Finally, character-level segmentation and Morfessor do not allow
to 
determine
the vocabulary size.
%We report the resulting dictionaries in Table \ref{tab:dictsize}.
Table \ref{tab:dictsize} shows the size of the resulting dictionaries.

\subsection{Evaluation}
\label{sec:evaluation}

%Motivation: We want a clean environment to test our hypothesis; as such we are not interested in building SOTA but allow for comparisons → line of interpretability. 
%N.B. the amount of data is variable for our 3 language pairs → not huge differences + we compare models trained on the same corpus.
We are interested in measuring both 
\textit{i)} the
overall translation quality obtained by different segmentation techniques, and \textit{ii)} the correct generation of gender forms.
%and the ability to produce the correct gender forms.
We evaluate 
%the
translation quality on both the MuST-C \textit{tst-COMMON} set (2,574 sentences for en-it and 2,632 for en-fr) and MuST-SHE ($\S$\ref{sec:preliminary}), using SacreBLEU~\cite{post-2018-a-call}.\footnote{
%The version signature is:
\texttt{BLEU+c.mixed+\#.1+s.exp+tok.13a+v.1.4.3}.}  

For fine-grained analysis on gender translation, we rely on gender accuracy~\cite{gaido-etal-2020-breeding}.\footnote{Evaluation script available with the MuST-SHE release.}
%\url{https://ict.fbk.eu/must-she/}.} 
We 
%also 
differentiate between two categories of phenomena represented in MuST-SHE. Category (1) contains first-person references (e.g. \textit{I'm a student}) to be translated according to the speakers' preferred linguistic expression of gender.
%\footnote{\bs{Speakers' \textit{he/she} pronouns have been manually retrieved.}} 
In this context, 
%direct
ST models can leverage speakers' vocal characteristics %and use it 
as a gender cue to infer gender translation.\footnote{Although they do not emerge in our experimental
%setting and evaluation set
settings, the potential risks of such capability are discussed in $\S$\ref{sect:ethic}.}
Gender phenomena of Category (2), instead, shall be translated in concordance with other gender information  
in
%within 
the sentence (e.g. \textit{\underline{she/he} is a student)}. 

% We measure translation quality with SacreBLEU\footnote{The version signature is: \texttt{BLEU+c.mixed+\#.1+s.exp+tok.13a+v.1.4.3}.} \cite{post-2018-a-call} and gender translation with gender accuracy \cite{gaido-etal-2020-breeding}.\footnote{Evaluation script available with the MuST-SHE release \url{https://ict.fbk.eu/must-she/}.}
% % \footnote{Computed with: \url{https://github.com/mgaido91/FBK-fairseq-ST/blob/master/scripts/eval/mustshe_acc.py}.}

%\cite{bentivogli-2020-gender}, a manually checked test set (1,096 sentences for en-it and 1,113 for en-fr) with annotation of gender-marked words to evaluate systems' ability to translate gender. MuST-SHE is divided into two main categories: 
%category 1 refers to words whose correct translation depends on the gender of the speaker, and category 2 to those in which the gender of the translated words is derived by a pronoun or another referent present in the sentence. 
%Each of the categories is further segmented into feminine and masculine forms.
%\lb{gender-balanced dev set tolto}

% So to avoid rewarding models' potentially biased behaviour, as a validation set we rely on the MuST-C gender-balanced dev set \cite{gaido-etal-2020-breeding}. 
%as we want to ensure our models are not biased by an unbalanced validation set.

\begin{table}[t]
\setlength{\tabcolsep}{4pt}
\centering
\small
\begin{tabular}{l|cc|c||cc|c}
& \multicolumn{3}{c||}{\textbf{en-fr}} & \multicolumn{3}{c}{\textbf{en-it}} \\
\hline
& M-C & M-SHE & Avg. & M-C & M-SHE & Avg. \\
\hline
BPE         & \textbf{30.7} & 25.9 & \textbf{28.3} & 21.4 & \textbf{21.8} & 21.6 \\
Char        & 29.5 & 24.2 & 26.9 & 21.3 & 20.7 & 21.0 \\
DPE         & 29.8 & 25.3 & 27.6 & 21.9 & 21.7 & \textbf{21.8} \\
Morfessor   & 29.7 & 25.7 & 27.7 & 21.7 & 21.4 & 21.6 \\
LMVR        & 30.3 & \textbf{26.0} & 28.2 & \textbf{22.0} & 21.5 & \textbf{21.8} \\
\end{tabular}
\caption{SacreBLEU scores on MuST-C tst-COMMON (M-C) and MuST-SHE (M-SHE) for 
%English-French and English-Italian.
en-fr and en-it.
}
  \label{tab:bleu}
\end{table}

\section{Comparison of Segmentation Methods}
\label{sec:comparison}
Table \ref{tab:bleu} shows the overall \textbf{translation quality} of ST systems trained with distinct segmentation techniques. 
%From the results, 
BPE comes out as competitive as LMVR 
for both language pairs. 
On averaged results, it
%It
%, but
exhibits a small gap
%deficit
(0.2 BLEU) also with DPE on en-it, while it achieves the best performance on en-fr.
%. On the French side instead, BPE achieves the best performance,
%\mn{closely followed by DPE}. 
The disparities are small though: they %all 
range within 0.5 BLEU, apart from Char standing $\sim$1 BLEU below. Compared to the scores reported by \citet{di-gangi-2020-target}, the Char gap is
%indeed
however smaller.  As our results are considerably higher than theirs, we believe that the reason for such differences lies in a sub-optimal fine-tuning of their hyper-parameters. Overall, in light of the trade-off between computational cost (LMVR and DPE require a dedicated training phase for data segmentation) and average performance (BPE achieves winning scores on en-fr and competitive for en-it), we hold BPE as the
%desired upper bound
best segmentation strategy in terms of general translation quality for direct ST.

Turning to \textbf{gender translation}, the gender accuracy scores presented in Table \ref{tab:genderacc} exhibit that all ST models are clearly biased, with masculine forms (M) disproportionately produced across language pairs and categories. 
%%%%%%%%%%%%% ORIG
% Turning now to \textbf{gender translation}, the gender accuracy scores (Table \ref{tab:genderacc}) appear consistent across language pairs, except for the comparatively hindered translation of French masculine forms of Category 1 (1M). Overall, they exhibit that all ST models are clearly biased, with masculine forms disproportionately produced across language pairs and categories \bs{(see Masculine (M) vs Feminine (F))}.
%%%%%%%%%%%%%
However, we intend to pinpoint the relative gains and losses among segmenting methods. 
%First, we attest that scores are consistent across language pairs, except for the relatively hindered translation of French masculine forms of Category 1 (1M). 
Focusing on overall accuracy (ALL), we see that
%Generally, 
Char -- despite its lowest performance in terms of BLEU score --
%overall translation quality -- 
emerges as the favourite segmentation 
%strategy
for gender translation. For French, however, 
DPE is only slightly behind. Looking at morphological methods, they surprisingly do not outperform the statistical ones.
The greatest variations are detected for feminine forms of Category 1 (1F), where none of the segmentation techniques reaches 50\% of accuracy, meaning that they are all worse than a random choice when the speaker should be addressed by feminine expressions.  Char appears close to such threshold, while the others (apart from DPE in French) are
significantly lower.
%slightly over 40\% or below. 

%Our results thus
These results illustrate that 
%\mg{the} 
target segmentation 
%\mn{technique} 
is a relevant parameter for gender translation. 
%%%
In particular, 
%\bs{for Category 1}, 
they suggest that Char segmentation improves models'
ability to learn correlations between the 
received 
%audio 
input and
%patterns in the 
%associated 
%data.
%the 
gender forms in the 
%target
\lb{reference}
translations.
%
%%%%%%%%%% nuovo
Although in this experiment models rely only on speakers' vocal characteristics 
%\bs{as a correlation to infer gender}
to infer gender -- 
which we discourage as a cue for gender translation for real-world deployment (see {$\S$}\ref{sect:ethic}) -- such ability shows a potential advantage for Char,
\lb{which}
\bs{could be better redirected toward learning 
%\mg{the correlation between the generated forms and}
\lb{correlations with}
reliable gender meta-information included in the input.}
%
%NOTE DA REBUTALL
%Our results suggest that Char better learns correlations from the audio input as it outperforms BPE on category1, where it infers speakers’ gender from their vocal characteristics. Since exploiting such information as a gender cue can be harmful to some users, we ponder that Char’s above-mentioned ability could be better redirected toward learning reliable gender meta-information included in the input. It is a future-work hypothesis (as we will better clarify in the camera ready) that we did not explore as we aimed to delve deeper into the implications of word segmentation.
%
%
%\bs{to be redirected for other scenarios.}
%\bs{which might be redirected toward more reliable inputs.}
For instance, in a scenario in which meta-information (e.g. a gender tag) is added to the input  to support gender translation, 
%\bs{it could be tested whether a}
a Char model 
%would 
might
better exploit this information.
%%%VECCHIO
%especially for generating. 
%In particular, it becomes a key factor to acquire
%the correlation between the speaker’s audio input and the corresponding linguistic form to be rendered in the translation \mg{(we do not advocate that models should be based on this correlation, but this capability of the models is significant as it indicates they are able to autonomously learn the correct gender to produce if this information is present in the input).}
%
%%
%ULTIMO DI MARCO 30.01
%\mg{In particular, it enables or prevents models from learning the correlation between the speaker’s audio input and the corresponding linguistic gender form to be rendered in the translation. Although we do not advocate that models should rely on physical cues to determine the gender to be rendered in the translation, this ability shows the potential of the technique to extract this kind of information from the input (in real word applications, the speaker's preferred gender might be an additional input for the model).}
%
%, [\BS\textbf{{A POTENTIALITY THAT COULD BE USEFUL ALSO PER? MARCOG IDEE PER FARE NUANCE?]}}}.
Lastly, our evaluation reveals that,
%\mg{, 
unlike %done in 
previous ST
%work 
studies \cite{bentivogli-2020-gender}, a 
%fair
proper
comparison of models' gender translation potentialities requires 
%the adoption of 
adopting the same segmentation.
%, unlike previous studies in ST \cite{bentivogli-2020-gender}, models should not be trained by means of the different segmentation technique to properly compare their gender translation potentialities.
%
%[DA RIVEDERE \mg{in the ability of a model to learn the correlation between the speaker’s vocal characteristics and the gender to be rendered in the translation. Moreover, models trained with different segmentation techniques cannot be compared in terms of their gender bias, as it was instead done in previous studies (Bentivogli et al., 2020).}
%
%
Our question then becomes: What makes Char segmentation less biased? 
%than the others?
 What are the 
%splitting's 
tokenization features %peculiarities
%Our analysis on the effect of different segmentation techniques on gender bias continues with the attempt to find the peculiarities
determining a better/worse ability in generating the correct gender forms?
\begin{table}[t]
\centering
\small
\begin{tabular}{l|c|cccc}
& \multicolumn{5}{c}{\textbf{en-fr}}  \\
\hline
& ALL & 1F & 1M & 2F & 2M \\%& ALL & 1F & 1M & 2F & 2M \\
\hline
BPE         & 65.18 & 37.17 & 75.44 & 61.20 & 80.80  \\
Char        & \textbf{68.85} & 48.21 & 74.78 & 65.89 & \textbf{81.03}  \\
DPE         & 68.55 & \textbf{49.12} & 70.29 & \textbf{66.22} & 80.90  \\
Morfessor   & 67.05 & 42.73 & 75.11 & 63.02 & 80.98  \\
LMVR        & 65.38 & 32.89 & \textbf{76.96} & 61.87 & 79.95 \\
\hline \hline
& \multicolumn{5}{c}{\textbf{en-it}} \\
\hline
BPE         &  67.47 & 33.17 & 88.50 & 60.26 & 81.82 \\
Char        &  \textbf{71.69} & \textbf{48.33} & 85.07 & \textbf{64.65} & \textbf{84.33} \\
DPE         &  68.86 & 44.83 & 81.58 & 59.32 & 82.62 \\
Morfessor   &  65.46 & 36.61 & 81.04 & 56.94 & 79.61 \\
LMVR        & 69.77 & 39.64 & \textbf{89.00} & 63.85 & 83.03 \\

\end{tabular}
\caption{Gender accuracy (\%) for MuST-SHE Overall (ALL), Category 1 and 2 on 
%English-French and English-Italian.
en-fr and en-it.
}
  \label{tab:genderacc}
\end{table}
\paragraph{Lexical diversity.}
%\mn{We conjecture if}
We posit that
the limited
%translation
generation of feminine forms can be framed as an issue of data sparsity, whereas the advantage of Char-based 
segmentation
%tokenization 
 ensues from its ability to handle less 
 %in
 frequent and unseen words \cite{Belinkov2019OnTL}. 
 Accordingly,
 %In fact, 
\citet{vanmassenhove-etal-2018-getting, roberts2020decoding} link the loss of \textit{linguistic diversity} (i.e. the range of lexical items used 
in a text) 
%and
with
%\lb{to}
the overfitted distribution of masculine references in MT outputs. 

To explore such hypothesis, we
compare the lexical diversity (LD)
%\footnote{We used: \url{https://github.com/LSYS/LexicalRichness/blob/master/lexicalrichness/lexicalrichness.py}.}
of our models' translations
%to
and MuST-SHE references. 
 To this aim, we rely on
 %\textit{i)} Type/Token ratio \textbf{(TTR)} \cite{chotlos1944iv, templin1957certain}, a simple and yet often used metric, but potentially affected by fluctuations due to text length \cite{tweedie1998variable}; and
% % %;
%\textit{ii}) the more robust Moving Average Type/Token ratio \textbf{(MATTR)} \cite{covington2010cutting},  which first calculates TTRs for successive non-overlapping segments of tokens within a given length (window size), and then provides the mean value of the estimated TTRs. 
Type/Token ratio \textbf{(TTR)} -- \cite{chotlos1944iv, templin1957certain}, and the more robust Moving Average TTR \textbf{(MATTR)} -- \cite{covington2010cutting}.\footnote{Metrics computed with software available at: \url{https://github.com/LSYS/LexicalRichness}. We set 1,000 as \texttt{window\_size} 
  for MATTR.}
  %\footnote{Metrics computed with: \url{https://github.com/LSYS/LexicalRichness/blob/master/lexicalrichness/lexicalrichness.py}. We set 1,000 as \texttt{window\_size}  for MATTR.}
%

As we can see in Table \ref{tab:lexicaldiv}, character-based models exhibit the highest
LD (the only exception is DPE with the less reliable TTR metric on en-it).
However, we cannot corroborate the hypothesis formulated in 
%previous works, 
the above-cited
%works
studies, as
%our 
LD scores do not strictly correlate with
%the gender accuracy scores
gender accuracy
(Table \ref{tab:genderacc}). For
instance, LMVR is the second-best in terms of gender accuracy on en-it, but shows a very low lexical diversity (the worst according to MATTR and second-worst according to TTR).

\begin{table}[h]
\centering
\small
\begin{tabular}{l|cc|cc}
& \multicolumn{2}{c|}{\textbf{en-fr}} & \multicolumn{2}{c}{\textbf{en-it}} \\
\hline
  & TTR & MATTR  & TTR & MATTR  \\
\hline
\textit{M-SHE Ref} 
        & \textit{16.12} &	\textit{41.39}& \textit{19.11} &	\textit{46.36} \\
BPE     & 14.53	& 39.69 & 17.46 & 44.86 \\
Char    & \textbf{14.97} &	\textbf{40.60} & 17.75 & \textbf{45.65} \\
DPE     & 14.83 & 40.02 & \textbf{18.07} & 45.12 \\
Morf    & 14.38 & 39.88 & 16.31 & 44.90 \\
LMVR    & 13.87 & 39.98 & 16.33 & 44.71
\end{tabular}
\caption{Lexical diversity scores on en-fr and en-it} 
%\mg{according to Type/Token ratio \textbf{(TTR)} \cite{chotlos1944iv, templin1957certain} andthe more robust Moving Average Type/Token ratio \textbf{(MATTR)} \cite{covington2010cutting}, with 1,000 as \texttt{window\_size}.}}
  \label{tab:lexicaldiv}
\end{table}

\paragraph{Sequence length.} 
As discussed in 
%section
$\S$\ref{sec:length}, we know that Italian and French feminine forms are, although to different extent, longer and less frequent than their masculine counterparts.
%comparatively longer than masculine ones and less frequent. 
In light of such conditions, we expected that the statistically-driven BPE segmentation would leave feminine forms unmerged at a higher rate, and thus add uncertainty to their generation.
%by our models.
To verify if this is the actual case --
%,
explaining BPE's lower gender accuracy
--
we  
%\bs{inspect the amount of fragments (sequence length) into which a segmented feminine word results with respect to its masculine pair.}
check whether the
% length of  the token (character or subword) sequence
number of tokens (characters or subwords) of a segmented feminine word
%differs with respect to its
is higher than that of the corresponding
 masculine form.
%
%
%Thus, we
We
%\bs{asses such ratio by exploting}
exploit 
the coupled 
%wrong and correct MuST-SHE references
``wrong'' and ``correct'' references available in MuST-SHE, and
compute the average percentage of additional 
tokens 
%\bs{fragments} 
found
in the
%feminine-sided
%feminine sentences
feminine segmented 
%words
sentences\footnote{As such references only vary for gender-marked words, we can isolate the difference relative to gender tokens.}
over the masculine ones. Results are reported in Table \ref{tab:lengths}.

\begin{table}[t]
\setlength{\tabcolsep}{4pt}
\centering
\small
\begin{tabular}{l|cc}
 & \textbf{en-fr} (\%) & \textbf{en-it} (\%) \\
\hline
BPE         & 1.04 & 0.88 \\
Char        & 1.37 & 0.38 \\
DPE         & 2.11 & 0.77 \\
Morfessor   & 1.62 & 0.45 \\
LMVR        & 1.43 & 0.33 \\
\hline
\end{tabular}
\caption{
%P ercentage (\%) of additional tokens that feminine sentences have compared to the masculine version of the same sentences in MuST-SHE. \mn{BRUTTA CAPTION.}
Percentage increase of token sequence's length
for feminine words over masculine ones.
%Percentual length difference between feminine/masculine pairs' sequences of \bs{word fragments}%tokens. 
%in terms of lengths (computed as the number of tokens) between feminine and masculine versions of the same sentences in MuST-SHE.
% Cioe', How longer feminine versions of tokenized sentences are than masculine ones
}
  \label{tab:lengths}
\end{table}

%we hypothesize that linguistically-unmotivated BPE segmentation would likely leave feminine forms unmerged at a higher \mn{rate, adding}  uncertainty to their generation \mn{by} our models.
%Table \ref{tab:lengths} reports the average percentage of additional tokens in the feminine version of the MuST-SHE sentences.\footnote{As MuST-SHE contains both the correct and the wrong reference, the masculine and feminine version of each sentence are available.}
At a first look, we 
observe opposite trends: 
BPE segmentation leads to the highest  increment of tokens for feminine words in Italian, but to the lowest one in French.
% while BPE shows the highest \bs{fragments} increment on %en-it
% Italian feminine forms, 
%it has the lowest one in French. 
Also, DPE exhibits the
highest 
%per cent increment of number of tokens in French
increment in French, whereas it actually performs slightly better than Char on feminine gender translation (see Table \ref{tab:genderacc}). Hence, even the increase in sequence length does not seem to be an issue on its own for gender translation. Nonetheless, 
%although the amount of splitting for feminine forms does not appear significant \textit{per se}, 
these apparently contradictory results encourage our last exploration: \textit{How} are gender forms actually split?
%.

\paragraph{Gender isolation.}
%By means of further manual analysis, we consider the different characteristics and structural mechanisms of the two target languages to mutually illustrate word segmentation’s tokens increment and gender translation ability.  

By means of further manual analysis \bs{on 50 output sentences per each of the 6 systems,} 
we 
%investigate
inquire
if longer
%tokens'
token sequences 
for feminine words
% segmentation's 
% %performance and
% %asymmetrical 
% fragments increment 
can be explained
%illustrated
in light of the different 
characteristics and gender productive mechanisms of the two target languages ({$\S$}\ref{sec:language}). Table \ref{tab:samples_segm} reports selected instances of coupled
%masculine/feminine
 feminine/masculine segmented words, 
%accompanied by
 with their respective frequency in the 
MuST-C training set.

% SECTION 3.1., as presented in Section {$\S$}\ref{sec:language}, where we showed that feminine variants are less frequent than their masculine counterparts in over 86\% of the cases.
% Looking at words' length, 15\% of Italian feminine forms result to be longer than masculine ones, 
% whereas in French this percentage amounts to almost 95\%. 
%
%In Italian, BPE increment indeed ensues from greedy splitting, which ignores meaningful affix boundaries both for same length (Table \ref{tab:samples_segm} example a) and different-length pairs (Table \ref{tab:samples_segm}.b). 
%Conversely, on the French set -- with 95\% of %its feminine entries built on more characters than their masculine counterparts -- BPE low increment is precisely due to  its  loss of semantic units. 

Starting with Italian, we find that BPE sequence length increment indeed ensues from greedy splitting that, as we can see from 
%Table \ref{tab:samples_segm} 
examples \textbf{(a)} and \textbf{(c)}, ignores meaningful affix boundaries for both same length and different-length gender pairs, respectively.
Conversely, on the French set – with 95\% of feminine words longer
%entries built on more characters 
than their masculine counterparts – 
%BPE
BPE's
low increment is precisely due to  its  loss  of  semantic  units. 
%In fact
For instance, as shown in \textbf{(e)}, BPE does not preserve the verb root (\textit{adopt}), nor %does it 
isolates the additional
%fragment
token (\textit{-e}) responsible for the feminine form, thus resulting
%in
into two words with the same sequence length (2 tokens). Instead DPE, which achieved the highest accuracy results for en-fr feminine translation (Table \ref{tab:genderacc}), treats the feminine additional character as a
%fragment
token \textit{per se} (\textbf{f}).
%
% As Table \ref{tab:samples_segm}.c shows, BPE does not preserve the verb root (\textit{adopt}), nor does it isolate the additional token (\textit{e}) responsible for the feminine form. DPE, instead,
% treats the feminine additional character as a token \textit{per se} (Table \ref{tab:samples_segm}.d).
% Thus, our intuition is that the proper splitting of the morpheme-econded gender information as a distinct meaningful token may favour gender translation, as models learn to productively generalize on it.  Considering the high increment of DPE tokens for Italian for the limited number of longer feminine forms (15\%), DPE is unlikely to isolate it on the en-it language pair. Its tokens increment is in fact similar to that of BPE, producing the same types of splitting (see Table \ref{tab:samples_segm}.a and b).
%

Based on such patterns, our intuition is that the proper splitting of the morpheme-encoded gender information as a distinct 
%meaningful
%fragment
token 
%\bs{may favour} 
favours 
gender translation, as models learn to productively generalize on it.  Considering the high increment of DPE
%fragments
tokens for Italian in spite of the limited number of longer feminine forms (15\%), our analysis confirms that DPE is unlikely to
%does not
isolate  gender morphemes %for en-it.
on the en-it language pair. 
As a matter of fact, it produces the same kind of coarse splitting as BPE (see \textbf{(b)} and \textbf{(d)}).
\begin{table}[t]
\centering
\setlength{\tabcolsep}{2.5pt}
\small
\begin{tabular}{lllllr}
& \textbf{en} & \textbf{Segm.} & \textbf{F} & \textbf{M} & \textbf{Freq. F/M} \\
\hline
a) & asked & BPE & chie--st\underline{a} & chiest\underline{o} & 36/884 \\
b) &  & DPE & chie--st\underline{a} & chiest\underline{o} & 36/884 \\
c) & friends & BPE & a--mic\underline{he} & amic\underline{i} & 49/1094 \\
d) &  & DPE & a--mic\underline{he} & amic\underline{i} & 49/1094 \\
e) & adopted & BPE & adop--té\underline{e} & adop--té & 30/103 \\
f) & & DPE & adop--t--é--\underline{e} & adop--t--é & 30/103 \\
\hline
g) & sure & Morf. & si--cur\underline{a} & sicur\underline{o} & 258/818 \\
h) & grown up & LMVR & cresci--ut\underline{a} & cresci--ut\underline{o} & 229/272 \\
i) & celebrated & LMVR & célébr--é\underline{e}s & célébr--és & 3/7
\end{tabular}
\caption{Examples of word segmentation. The segmentation boundary is identified by "--".}
  \label{tab:samples_segm}
\end{table}

%
%%%PRECEDENTE
%Backed up by further manual analysis, we explain \mg{the different abilities to translate gender} in light of the different characteristics and linguistic productive  \mn{mechanisms} of the two \mn{target languages.}
%In Italian,  BPE increment is indeed often  the result of greedy splitting that ignores meaningful affix boundaries both for same length (Table \ref{tab:samples_segm}.a) as well as for different-length pairs (Table \ref{tab:samples_segm}.b).
%
%\footnote{Tokens frequency: amiche(49) - amici(1094), chiesta(36) - chiesto(884) }}
%
%\
%\textbf{(a)} asked: F. chie@st\underline{a} \textit{vs} M. chiest\underline{o}\
%
%\textbf{(b)} friends: a@mic\underline{he} F. \textit{vs} amic\underline{i} M. %
%
%\bs{Conversely, with the French set having 95\% of its feminine entries containing more characters than their masculine counterparts, its low increment is precisely due to its inability to preserve semantic units. 
%
%Therefore, our intuition is that as long as the gender information is properly preserved in the splitting as a distinct meaningful token, gender translation may be favoured as models learn to productively generalize on such information. Considering the high increment of DPE tokens for Italian for the limited number of longer feminine forms
%in the Italian set (15\%)}, DPE is unlikely to \mg{achieve} it on the en-it language pair. Its tokens increment is in fact similar to that of BPE,
%\mg{and it produces the same types of splitting of BPE (see Table \ref{tab:samples_segm}.a and b).}

Finally, we attest that the two morphological techniques are not equally valid. 
%\bs{Rather},
Morfessor occasionally generates morphologically incorrect subwords for feminine forms by breaking the word stem (see example \textbf{(g)} where the correct stem is \textit{sicur}).
%\bs{by breaking the word stem (\textit{sicur}) as in \textbf{(e)}}. 
%\bs{Thus}, 
Such behavior also explains Morfessor’s higher token increment with respect to LMVR.  Instead, although LMVR (examples \textbf{(h)} and \textbf{(i)})  produces 
linguistically valid suffixes,
%semantically meaningful suffixes, 
it often condenses other grammatical categories (e.g. tense and number) %together 
with gender.
%(Table \ref{tab:samples_segm}.f and g). 
As suggested above, if the
pinpointed split
%\mg{isolation} 
of morpheme-encoded gender is a key factor for gender translation, 	LMVR’s %higher 
lower level of granularity %may 
%\bs{would explain}
explains 
its %lower 
reduced
gender accuracy.
%performance.
Working on character' sequences, instead, the isolation of the %single 
gender unit is always
%at reach
attained. 

\section{Beyond the Quality-Gender Trade-off}

Informed by 
our experiments and analysis
%--
%%analysis 
%and given the computational factor discussed 
%\mg{our experiments and analysis} 
%in 
({$\S$}\ref{sec:comparison})\mg{,}
%--
%%and informed by the conducted analysis,
%%Informed by the so far conducted experiments and analysis, 
%
%we pursue the creation of a model
%
we conclude this study by proposing a model
that combines BPE overall translation quality and Char's ability to translate gender. 
%Such a system would overcome the trade-off between generic performance and gender representation.
%Specifically, 
%Accordingly, 
To this aim, we 
%propose
train
a multi-decoder approach that exploits both 
%the 
segmentations to draw on their corresponding advantages. 
%extract the best the model could learn with each of them.

In the context of ST, several multi-decoder architectures have been proposed, usually to jointly produce both transcripts and translations with a single model. Among those in which both decoders access the encoder output, %in this study 
here we consider
%ed 
the best performing architectures according to \citet{sperber-et-al-2020-consistent}. 
%\mn{among those in which both decoders access the encoder output.} 
As such, we consider:  \textit{i)} 
%the 
\textit{Multitask direct}, a model with one encoder and two decoders, both exclusively attending the encoder output 
as proposed by \citet{weiss2017sequence}, and \textit{ii)} the \textit{Triangle} model \cite{anastasopoulos-2018-multitask}, 
in which the 
%\bs{with a} 
second decoder 
%\mt{\st{can}} 
%\bs{that}
attends
%both 
the output
%layer
of both the encoder and the first decoder.

For the triangle model, we 
used a first BPE-based decoder and a second Char decoder. With this order, we aimed to enrich BPE high quality translation 
%Our goal was to enrich the high quality translation {--} produced by BPE --
with a refinement for gender translation,
performed by the Char-based decoder. However, the results were negative:
the second decoder seems to
excessively rely 
on the output of the first one, thus suffering from a severe
\textit{exposure bias} \cite{ranzato2016sequence} at inference time. 
%For this reason 
%As such,
%Thus,
Hence, we do not report the results of 
these experiments.
%, but we leave to future works the study of methods to alleviate the \textit{exposure bias} issue.
%
%
%
%Since the Char results are not significantly enhanced with respect to the base Char encoder-decoder model, we only report the {overall translation quality} (Table \ref{tab:bleu_bpechar}) and {gender accuracy} (Table \ref{tab:genderacc_bpechar}) \mg{of the BPE output (\textbf{BPE}\&Char)}.
%
%
%For the \textit{Multitask direct} (\texttt{BPE\&Char} in Table \ref{tab:bleu_bpechar} and \ref{tab:genderacc_bpechar}), instead, we report the results of the BPE decoder, as the translation quality of the Char decoder is not significantly enhanced with respect to the base Char encoder-decoder model, while the gender accuracy of the BPE-based decoder is actually improved by the presence of the Char-based decoder.
%
%Considering the former, we attest that our proposed model achieves the best average performance. 
\begin{table}[t]
\setlength{\tabcolsep}{3pt}
\centering
\small
\begin{tabular}{l|cc|c||cc|c}
& \multicolumn{3}{c||}{\textbf{en-fr}} & \multicolumn{3}{c}{\textbf{en-it}} \\
\hline
& M-C & M-SHE & Avg. & M-C & M-SHE & Avg. \\
\hline
BPE         & \textbf{30.7} & 25.9 & 28.3 & 21.4 & 21.8 & 21.6 \\
Char        & 29.5 & 24.2 & 26.9 & 21.3 & 20.7 & 21.0 \\
\hline
\hline
BPE\&Char & 30.4 & \textbf{26.5} & \textbf{28.5} & \textbf{22.1} & \textbf{22.6} & \textbf{22.3} \\
\end{tabular}
\caption{SacreBLEU scores on MuST-C tst-COMMON (M-C) and MuST-SHE (M-SHE) on en-fr and en-it.}% English-French and English-Italian.}
  \label{tab:bleu_bpechar}
\end{table}

\begin{table}[t]
\centering
\small
\begin{tabular}{l|c|cccc}
& \multicolumn{5}{c}{\textbf{en-fr}}\\% & \multicolumn{5}{c}{\textbf{en-it}} \\
\hline
& ALL & 1F & 1M & 2F & 2M \\
\hline
BPE         & 65.18 & 37.17 & \textbf{75.44} & 61.20 & 80.80  \\
Char        & \textbf{68.85} & \textbf{48.21} & 74.78 & 65.89 & 81.03   \\
\hline
\hline
BPE\&Char & 68.04 & 40.61 & 75.11 & \textbf{67.01} & \textbf{81.45}  \\
%\hline
%\hline
\hline
& \multicolumn{5}{c}{\textbf{en-it}} \\
\hline
BPE         & 67.47 & 33.17 & \textbf{88.50} & 60.26 & 81.82 \\
Char        &  \textbf{71.69} & 48.33 & 85.07 & \textbf{64.65} & \textbf{84.33} \\
\hline
\hline
BPE\&Char &  70.05 & \textbf{52.23} & 84.19 & 59.60 & 81.37 \\
\end{tabular}
\caption{Gender accuracy (\%) for MuST-SHE Overall (ALL), Category 1 and 2 on en-fr and en-it.} %English-French and English-Italian.}
  \label{tab:genderacc_bpechar}
\end{table}
Instead, the \textit{Multitask direct} has one BPE-based and one Char-based decoder.
The system
%requires only
requires a training time increase of only 10\% and 20\% compared to, respectively, Char and BPE models. At inference phase, instead,
%instead, 
running time and size
%remain unvaried.
are the same of a BPE model.
%in the inference phase.} 
%
%\bs{requiring only a 10\% and 20\% training time increase over Char and BPE, respectively.}
%
We report overall translation quality (Table \ref{tab:bleu_bpechar}) and gender accuracy (Table \ref{tab:genderacc_bpechar}) of the BPE output (BPE\&Char).\footnote{The Char scores are not reported, as they are not %significantly 
enhanced compared to the base Char encoder-decoder model.}
Starting with gender accuracy,
%results show that 
the Multitask model's overall gender translation ability (ALL) is still lower, although very close, to that of the Char-based model.
Nevertheless, feminine translation improvements are 
%distributed are 
%in fact 
present on Category 2F for en-fr and, with a larger gain, on 1F for en-it.
%Category 1F is the one that benefits most and the overall ability to translate gender is close, even though still lower, to that of the Char-based model.
We believe that the presence of the Char-based decoder is beneficial to
%helps by learning to 
capture into %the gender information 
the encoder output %the 
gender information,
%that 
which
is then also 
exploited by the BPE-based decoder.
%\bs{, too.}
As
%more information is present in the encoder outputs,
the encoder outputs  are richer, overall translation quality is
%By extracting more information in the encoder output, the translation quality is 
also slightly improved (Table \ref{tab:bleu_bpechar}).
%\bs{Such achievements require only a training time increase of 10\% and 20\% compared to, respectively, Char and BPE models. The multitask's running time and size remain unvaried in the inference phase.} 
%\mg{These improvements come at the cost of an increased training time by 10\% with respect to the Char model and 20\% to the BPE model. At inference time, instead, the models' running time and size is exactly the same of the BPE model.}
%
%\bs{Overall, this last set of experiments
%suggest that} segmentation techniques determine what a model is able to extract from the input audio, thus also changing \bs{the information that is represented in the encoder output.} 
%
%Bea: speriamo sia quella buona
%%%
%\bs{Following these considerations, it appears that the conventional vision separating the two components’ activity in encoder-decoder architectures does not apply. In fact, although the 
%%Char and BPE
%two
%decoders do not interact with one another in the Multitask model, the system is nonetheless benefiting from Char’s attested ability to handle gender translation. As such, the chosen segmentation technique does not merely seem to determine a model’s improved capacity to extract %source 
%gender information in the encoder, and then render it into the target language. Rather, the target word segmentation might directly influence the gender information expressed and retained in the input’s internal representation, which is then leveraged in translation.} 
%
%
%
%
This finding is in line with other work \cite{costajussa2020gender}, which proved a strict relation between 
gender accuracy and the amount of gender information 
%encoded
retained
in the intermediate representations (encoder outputs).  

\lb{Overall, following these considerations, we posit that target segmentation can directly influence the gender information captured in the encoder output. In fact, since the Char and BPE decoders do not interact with each other in the Multitask model, the gender accuracy gains of the BPE decoder cannot be attributed to a better ability of a 
%certain 
segmentation method in rendering the gender information present in the encoder output into the translation.}

Our results
%paves
pave
the way for future %lines of 
research %into 
on
the creation of richer encoder outputs, disclosing
%the potential advantage of multiple segmentations to face
the importance of %the 
target segmentation
in extracting gender-related knowledge.
%the gender bias’ shortfall of current translation models.
With this work, we have taken a step forward in ST \bs{for English-French and English-Italian},
pointing at plenty of new 
%exciting 
ground to cover concerning \textit{how to} split
%.
%. 
\bs{for different language typologies.}
As the motivations 
%question 
of this inquiry clearly concern MT as well, we invite novel studies %are thus invited 
to start from our discoveries and explore 
\bs{how they apply under such conditions, as well as their combination with other bias mitigating strategies.}
\label{sect:pdf}
%same word translation varies according to context → context is relevant for gender translation, not only single word
%attested different segmentation between same length masculine/feminine forms
%patterns of which some connoted words that are always and across segmentation mis-translated
%some syntactical construction or distribution of gender cues (cat.2) more complex to translate. 
%
\section{Conclusion}

As the old IT saying goes: \textit{garbage in, garbage out}. This assumption underlies most of current attempts to address gender bias in language technologies. 
%models, ST included. 
%Rather then focusing on the well-know source of gender imbalance in the training data, 
Instead, in this work we explored whether technical choices can exacerbate gender bias by focusing on the influence of
%Specifically, we inspected if a models’ 
%critical component such as 
word segmentation 
%might affect 
on gender translation in ST. To this aim, we compared several word segmentation approaches on the target side of ST systems for English-French and English-Italian, in light of the linguistic gender features of the two 
%output grammatical gender 
target languages.  Our results show that tokenization does affect gender translation, 
%but also 
and that the higher BLEU scores of state-of-the-art BPE-based models come at cost of lower gender accuracy.  Moreover, \bs{first} analyses on the behaviour of segmentation techniques found that improved generation of gender forms 
\bs{could be}
%is 
linked to the proper isolation of the morpheme that encodes gender information, a feature which is attained by character-level split.
%\mn{unit}.
%splitting. 
%Accordingly,
Finally,
%informed by our analysis and experiments, 
we propose a multi-decoder approach to leverage the qualities of both BPE and character splitting, 
%Overall, such model exhibit a desidered trade-off, %between overall quality and gender recognition, 
improving both 
gender accuracy and BLEU score,  
%with a small gain in, 
while keeping computational costs under control. 

\section*{Acknowledgments}

This work is part of the ``End-to-end Spoken Language Translation in Rich Data Conditions'' project,\footnote{\url{https://ict.fbk.eu/units-hlt-mt-e2eslt/}} which is financially supported by an Amazon AWS ML Grant. The authors also wish to thank Duygu Ataman for the insightful discussions on this work.

\section{Ethic statement\footnote{Extra space after the 8th page allowed for ethical considerations -- see \url{https://2021.aclweb.org/calls/papers/}}}
\label{sect:ethic}

In compliance with ACL norms of ethics, we wish to %further 
elaborate on \textit{i)} characteristics of the dataset used in our experiments, \textit{ii)} the  study of  gender as a variable, and \textit{iii)} the %unintended 
harms potentially arising from real-word deployment of direct ST technology. 

As already stated, in our experiments we rely on the training data from the TED-based MuST-C corpus\footnote{\url{https://ict.fbk.eu/must-c/}}
%training data 
and its derived evaluation benchmark, MuST-SHE. Although precise information about various sociodemographic groups represented in the data are not fully available, based on impressionistic overview and prior knowledge about the nature of TED talks it is expected that the speakers are almost exclusively adults (over 20), with different 
%cultural
geographical
backgrounds. Thus, such data are likely to allow for modeling 
%of 
a range of English varieties of both native and non-native speakers.

As regards gender, from the %available 
data statements \cite{bender-friedman-2018-data} of the used corpora, we know that MuST-C training data are manually annotated with speakers’ gender information\footnote{\url{https://ict.fbk.eu/must-speakers/}} based on the personal pronouns found in their publicly available personal TED profile. As reported in its release page,\footnote{\url{https://ict.fbk.eu/must-she/}} the same annotation process applies to MuST-SHE as well, with the additional check that the indicated (English) linguistic gender forms are rendered in the gold standard translations. Hence, information about %the
speakers' preferred linguistic expressions of gender are transparently validated and disclosed. Overall, MuST-C exhibits a gender imbalance: 70\% vs. 30\% of the speakers referred by means of \textit{he/she} pronoun, respectively. Instead, 
allowing for a proper cross-gender comparison, they are equally distributed in MuST-SHE.
%Full disclosure and details about both datasets and annotations are reported in the data statements \cite{bender-friedman-2018-data}
%at their release page \footnote{\url{https://ict.fbk.eu/must-she/} and \url{https://ict.fbk.eu/must-speakers/}}.

%As regards gender, MuST-C training data are manually annotated  with speakers’ gender information based on the personal pronouns found in their publicly available personal TED section. The same process applies to MuST-SHE as well, with the additional check that the indicated linguistic gender forms are rendered in the gold standard translation. Thus, information about the speaker’s preferred linguistic expressions of gender are transparently validated and available for such corpora. Full disclosure and details abut both datasets and annotations are reported in the data statements \cite{bender-friedman-2018-data} at their release page \footnote{\url{https://ict.fbk.eu/must-she/} and \url{https://ict.fbk.eu/must-speakers/}}. 

Accordingly, when working on the evaluation of speaker-related gender translation for MuST-SHE category (1), we proceed by solely focusing on the rendering of their %\bs{approved} 
linguistic gender expressions.  
%such desired gender expressions.
As per \cite{larson-2017gender} guidelines, no assumptions about speakers’ self determined identity \cite{GLAAD} -- which cannot be directly mapped from pronoun usage \cite{cao-daume-iii-2020-toward, ackerman2019syntactic} --  has been made. Unfortunately, our experiments only account for the binary linguistic forms represented in the used data. To the best of our knowledge, ST natural language corpora  going beyond binarism do not yet exist,\footnote{In the whole MuST-C training set, only one speaker with \textit{they} pronouns is represented.} also due to the fact that gender-neutralization strategies are still object of debate and challenging to 
%be 
fully 
implement
in languages with grammatical gender \cite{gabriel2018neutralising,lessinger2020challenges}. Nonetheless, we support the rise of alternative neutral expressions for both languages \cite{shroy2016innovations, gheno2019femminili} and
point towards the 
%highlight the need to work toward the 
development of non-binary inclusive technology.

Lastly, we endorse the point made by \citet{gaido-etal-2020-breeding}. Namely, 
%that 
direct ST systems leveraging speaker’s vocal biometric features as a gender cue have the capability to bring real-world dangers, like the categorization of individuals by means of biological essentialist frameworks \cite{zimmantransgender}. This is particularly harmful to transgender individuals, as it can lead to misgendering \cite{stryker2008transgender} and diminish their personal identity. More generally, it can reduce gender to  stereotypical expectations about how masculine or feminine voices should sound. 
Note that, we do not advocate for the deployment of ST technologies \textit{as is}. Rather, we experimented with unmodified models for the sake of hypothesis testing without adding variability. However,  
our results suggest that, if certain word segmentation techniques better capture correlations from the received input, such capability could be exploited to redirect ST attention away from speakers' vocal characteristics by means of other information provided.

\bibliographystyle{acl_natbib}
\bibliography{acl2021}

\appendix

\section{Models}
\label{appsec:models}

\begin{table}[h]
\centering
\begin{tabular}{l|c|c}
& \textbf{en-fr} & \textbf{en-it} \\
\hline
\hline
BPE         & 60M & 60M \\
Char        & 52M & 52M \\
DPE         & 60M & 60M \\
Morfessor   & 79M & 76M \\
LMVR        & 74M & 72M \\
\hline
BPE\&Char    & 77M & 77M \\
\end{tabular}
\caption{Number of parameters (in millions). For BPE\&Char, the reported number is the total of the training parameters, but at inference time only one decoder is used, so the size is the same of BPE.}
  \label{tab:params}
\end{table}

Our direct ST models are built with the Fairseq library \cite{ott2019fairseq} and are based on the
 Transformer \cite{transformer}.
 They have 11 encoder and 4 decoder layers
 %\cite{potapczyk-przybysz-2020-srpols,gaido-etal-2020-end}.
 \mg{\cite{potapczyk-przybysz-2020-srpols}.} The encoder layers are preceded by 2 3x3 convolutional layers with 64 filters that reduce the input sequence length by a factor of 4
 and their attention weights are added a logarithmic distance penalty \cite{diGangi2019enhancing}.
 The models are optimized on label smoothed cross-entropy \cite{szegedy2016rethinking} -- the smoothing factor is 0.1 -- with Adam using $\beta_1$=0.9, $\beta_2$=0.98 and the learning rate is linearly increased during the warm-up phase (4k iterations) up to the maximum value $5\times 10^{-3}$, followed by decay with inverse square root policy. The dropout is set to 0.2 and each mini-batch consists of 8 sentences, while the update frequency is 8. The source audio
 %is represented by 40 features extracted from 25ms windows with 10ms slide and 
 is augmented with SpecAugment \cite{Park_2019,bahar-2019-specaugment} that is applied with probability 0.5 by masking two bands on the frequency axis (with 13 as maximum mask length) and two on the time axis (with 20 as maximum mask length).
 
 The systems are trained on MuST-C \cite{MuST-Cjournal}.
 We filtered from the training set all the samples whose audio length is higher than 20s.
 So to avoid rewarding models' potentially biased behaviour, as a validation set we rely on the MuST-C gender-balanced dev set \cite{gaido-etal-2020-breeding}. The target text was tokenized with Moses.\footnote{\url{https://github.com/moses-smt/mosesdecoder}}  We normalized audio per-speaker and extracted 40 features with 25ms windows sliding by 10ms with XNMT\footnote{\url{https://github.com/neulab/xnmt}} \cite{XNMT}.
 
 Trainings are stopped after 5 epochs without improvements on the validation loss and we average 5 checkpoints around the best on the validation set.
 They were performed on 8 K80 GPUs and lasted 2-3 days.

\end{document}

\appendix

\section{Models}

\begin{table}[h]
\centering
\begin{tabular}{l|c|c}
& \textbf{en-fr} & \textbf{en-it} \\
\hline
\hline
BPE         & 60M & 60M \\
Char        & 52M & 52M \\
DPE         & 60M & 60M \\
Morfessor   & 79M & 76M \\
LMVR        & 74M & 72M \\
\hline
BPE\&Char    & 77M & 77M \\
\end{tabular}
\caption{Number of parameters (in millions). For BPE\&Char, the reported number is the total of the training parameters, but at inference time only one decoder is used, so the size is the same of BPE.}
  \label{tab:params}
\end{table}

Our direct ST models are built with the Fairseq library \cite{ott2019fairseq} and are based on the
 Transformer \cite{transformer}.\footnote{We will release our code open source base upon paper acceptance.}
 They have 11 encoder and 4 decoder layers
 %\cite{potapczyk-przybysz-2020-srpols,gaido-etal-2020-end}.
 \mg{\cite{potapczyk-przybysz-2020-srpols}.} The encoder layers are preceded by 2 3x3 convolutional layers with 64 filters that reduce the input sequence length by a factor of 4
 and their attention weights are added a logarithmic distance penalty \cite{diGangi2019enhancing}.
 The models are optimized on label smoothed cross-entropy \cite{szegedy2016rethinking} -- the smoothing factor is 0.1 -- with Adam using $\beta_1$=0.9, $\beta_2$=0.98 and the learning rate is linearly increased during the warm-up phase (4k iterations) up to the maximum value $5\times 10^{-3}$, followed by decay with inverse square root policy. The dropout is set to 0.2 and each mini-batch consists of 8 sentences, while the update frequency is 8. The source audio
 %is represented by 40 features extracted from 25ms windows with 10ms slide and 
 is augmented with SpecAugment \cite{Park_2019,bahar-2019-specaugment} that is applied with probability 0.5 by masking two bands on the frequency axis (with 13 as maximum mask length) and two on the time axis (with 20 as maximum mask length).
 
 The systems are trained on MuST-C \cite{MuST-Cjournal}.
 We filtered from the training set all the samples whose audio length is higher than 20s.
 So to avoid rewarding models' potentially biased behaviour, as a validation set we rely on the MuST-C gender-balanced dev set \cite{gaido-etal-2020-breeding}. The target text was tokenized with Moses.\footnote{\url{https://github.com/moses-smt/mosesdecoder}}  We normalized audio per-speaker and extracted 40 features with 25ms windows sliding by 10ms with XNMT\footnote{\url{https://github.com/neulab/xnmt}} \cite{XNMT}.
 
 Trainings are stopped after 5 epochs without improvements on the validation loss and we average 5 checkpoints around the best on the validation set.
 They were performed on 8 K80 GPUs and lasted 2-3 days.

%Link
%Release upon acceptance

%Validation Performance

%As there is only one set (MuST-SHE, which is a benchmark, test set) with the annotation of gender marked words,
%we cannot evaluate gender accuracy on the validation set.
%Since the focus is the comparison among different systems and we already have two test sets on which we evaluate the overall translation quality we did not compute it on the validation set.

%No link to a downloadable version of the data
%No Expected validation performance, or the mean and variance as a function of the number of hyperparameter trials

\bibliographystyle{acl_natbib}
\bibliography{acl2021}